\documentclass[10pt,twocolumn,letterpaper]{article}

\usepackage[pagenumbers]{cvpr}   

\usepackage{cuted} 

\usepackage{currfile} 

\usepackage{caption} 
\definecolor{cvprblue}{rgb}{0.21,0.49,0.74}
\usepackage[pagebackref,breaklinks,colorlinks,allcolors=cvprblue]{hyperref}
\usepackage{booktabs, multirow,graphicx}
\usepackage[table]{xcolor}
\usepackage{tikz}
\usepackage[T1]{fontenc}
\usepackage{lmodern}
\definecolor{lppRow}{RGB}{218,229,239}

\usepackage{pifont}
\newcommand{\cmark}{\ding{51}}
\newcommand{\xmark}{\ding{55}}

\def\confName{CVPR}
\def\confYear{2027}

\title{CLASP: Complementary Layer Selection and Profiling for Zero-Shot OOD Detection using Vision-Language Models}

\author{
\resizebox{0.98\textwidth}{!}{%
    \mbox{
        Saahil Islam\textsuperscript{1} \quad
        Adarsh Bhandary Panambur\textsuperscript{1} \quad
        Teodora Spasojevic\textsuperscript{2} \quad
        Eduardo Castañeda\textsuperscript{3} \quad
        Andreas Maier\textsuperscript{1}
    }
}\\[5pt]
{\small
\textsuperscript{1}Pattern Recognition Lab, Friedrich-Alexander-Universität Erlangen-Nürnberg} \\
{\small 
\textsuperscript{2}Artificial Intelligence in Biomedical Engineering, Friedrich-Alexander-Universität Erlangen-Nürnberg} \\ 
{\small 
\textsuperscript{3}Siemens Healthineers, Erlangen, Germany
}
}

\begin{document}
\maketitle

\begin{abstract}
Vision-language models (VLMs) offer a promising approach to zero-shot out-of-distribution (OOD) detection in medical imaging. However, existing methods predominantly rely on final-layer confidence scores or entropy-guided selection of intermediate representations, potentially overlooking complementary information across visual layers and textual concepts. We investigate these limitations and show that low in-distribution (ID) confidence entropy does not necessarily translate into improved OOD discrimination. We propose \textbf{CLASP (Complementary Layer Selection and Profiling)}, an ID-only framework that combines structured layer--prompt responses with visual representations for OOD detection. CLASP employs Gaussian conditional mutual information to rank encoder layers, followed by anchored spectral selection to automatically determine a compact subset without a predefined layer count. The selected layers define a matched layer--prompt profile, which is combined with PCA-compressed visual embeddings and modeled using a single shrinkage-regularized Gaussian. OOD samples are identified through squared Mahalanobis distance without model fine-tuning, external OOD data, or test-time adaptation. Experiments across histopathology, chest X-ray, and brain MRI benchmarks using UniMedCLIP and BioMedCLIP demonstrate strong performance across diverse distribution shifts. On MIDOG near-OOD, CLASP achieves 87.54\% AUROC with UniMedCLIP, substantially outperforming confidence-based methods, while performance with BioMedCLIP remains competitive but below visual-only Mahalanobis modeling. Further analyses demonstrate the importance of preserving layer--prompt interactions and modeling cross-feature dependencies, while revealing shift-dependent contributions from visual and prompt representations. Despite remaining challenges under semantic shifts, CLASP provides a unified approach to OOD detection across heterogeneous medical imaging domains.
\end{abstract}

\section{Introduction}
\label{sec:introduction}

Deep learning has demonstrated considerable potential in medical image analysis, including disease classification, segmentation, and computer-aided diagnosis. However, its reliability remains challenged by distribution shifts arising from variations in imaging devices, acquisition protocols, and patient populations. Such shifts can lead to unreliable predictions, even when models exhibit high confidence \cite{zech2018variable,gutbrod2025openmibood}. Out-of-distribution (OOD) detection seeks to identify inputs that deviate from the in-distribution (ID) population before their predictions are used for clinical decision-making. Vision-language models (VLMs), particularly medical adaptations such as BioMedCLIP and UniMed-CLIP \cite{zhang2025biomedclip,khattak2024unimedclip}, offer a promising direction by enabling zero-shot OOD detection through image--text alignment without requiring additional task-specific classifier training.

\begin{figure*}[t]
    \centering
    \begin{minipage}[c]{0.38\textwidth}
        \centering
        \includegraphics[width=\linewidth]{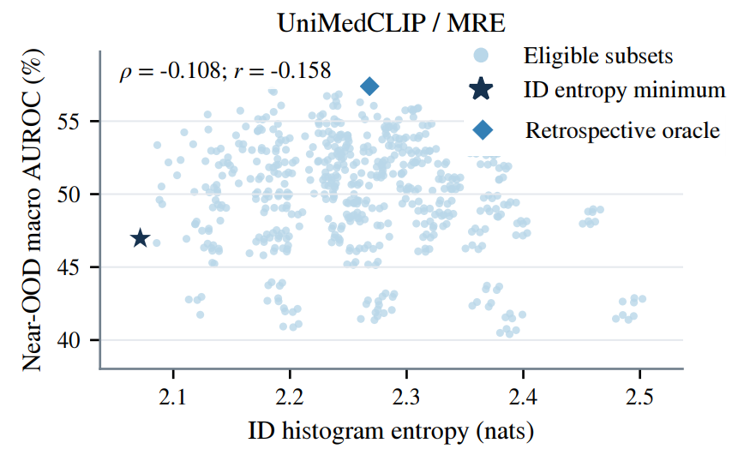}
        \vspace{2pt}
        {\small \textbf{(a)} Entropy-based layer selection}
    \end{minipage}
    \hfill
    \begin{minipage}[c]{0.60\textwidth}
        \centering
        \resizebox{\linewidth}{!}{\definecolor{titleblue}{RGB}{17,45,82}
\definecolor{labelblue}{RGB}{89,114,142}
\definecolor{shadeone}{RGB}{231,242,253}
\definecolor{shadetwo}{RGB}{201,226,248}
\definecolor{shadethree}{RGB}{166,208,241}
\definecolor{shadeseven}{RGB}{57,121,188}
\definecolor{shadeeight}{RGB}{35,103,173}
\definecolor{shadenine}{RGB}{17,81,145}
\definecolor{maxbg}{RGB}{237,245,255}

\newcommand{\heatcell}[5]{
  \fill[fill=#3,rounded corners=0.6mm]
    (#1,#2) rectangle ++(180,103);
  \node[
    font=\sffamily\bfseries\fontsize{12.5}{14.5}\selectfont,
    text=#4
  ] at ({#1+90},{#2+51.5}) {#5};
}

\newcommand{\maxcell}[3]{
  \fill[fill=maxbg,rounded corners=0.6mm]
    (#1,#2) rectangle ++(175,103);
  \node[
    font=\sffamily\bfseries\fontsize{12.5}{14.5}\selectfont,
    text=shadeeight
  ] at ({#1+87.5},{#2+51.5}) {#3};
}

\newcommand{\colhead}[2]{
  \node[
    font=\sffamily\bfseries\fontsize{10}{12}\selectfont,
    text=labelblue
  ] at (#1,132) {#2};
}

\newcommand{\rowhead}[3]{
  \node[
    font=\sffamily\bfseries\fontsize{11.5}{13.5}\selectfont,
    text=labelblue
  ] at (#1,#2) {#3};
}

\begin{tikzpicture}[
  x=0.1mm,
  y=-0.1mm,
  every node/.style={inner sep=0pt,outer sep=0pt}
]

\path[use as bounding box]
  (0,0) rectangle (1949,669);

\node[
  anchor=west,
  font=\sffamily\bfseries\fontsize{21}{24}\selectfont,
  text=titleblue
] at (91,55) {Image A};

\node[
  anchor=west,
  font=\sffamily\bfseries\fontsize{21}{24}\selectfont,
  text=titleblue
] at (1030,55) {Image B};

\colhead{198}{Prompt 1}
\colhead{404}{Prompt 2}
\colhead{609}{Prompt 3}
\colhead{875}{Max}

\colhead{1178}{Prompt 1}
\colhead{1383}{Prompt 2}
\colhead{1589}{Prompt 3}
\colhead{1855}{Max}

\foreach \y/\layer in {
  227/L1,
  354/L2,
  480/L3,
  607/L4
}{
  \rowhead{36}{\y}{\layer}
  \rowhead{1016}{\y}{\layer}
}

\heatcell{108}{175}{shadeeight}{white}{0.8}
\heatcell{314}{175}{shadetwo}{titleblue}{0.2}
\heatcell{519}{175}{shadeone}{titleblue}{0.1}
\maxcell{788}{175}{0.8}

\heatcell{108}{302}{shadeseven}{white}{0.7}
\heatcell{314}{302}{shadethree}{titleblue}{0.3}
\heatcell{519}{302}{shadeone}{titleblue}{0.1}
\maxcell{788}{302}{0.7}

\heatcell{108}{428}{shadenine}{white}{0.9}
\heatcell{314}{428}{shadetwo}{titleblue}{0.2}
\heatcell{519}{428}{shadeone}{titleblue}{0.1}
\maxcell{788}{428}{0.9}

\heatcell{108}{555}{shadeeight}{white}{0.8}
\heatcell{314}{555}{shadeone}{titleblue}{0.1}
\heatcell{519}{555}{shadethree}{titleblue}{0.3}
\maxcell{788}{555}{0.8}

\heatcell{1088}{175}{shadetwo}{titleblue}{0.2}
\heatcell{1293}{175}{shadeeight}{white}{0.8}
\heatcell{1499}{175}{shadeone}{titleblue}{0.1}
\maxcell{1767}{175}{0.8}

\heatcell{1088}{302}{shadeone}{titleblue}{0.1}
\heatcell{1293}{302}{shadethree}{titleblue}{0.3}
\heatcell{1499}{302}{shadeseven}{white}{0.7}
\maxcell{1767}{302}{0.7}

\heatcell{1088}{428}{shadeone}{titleblue}{0.1}
\heatcell{1293}{428}{shadenine}{white}{0.9}
\heatcell{1499}{428}{shadetwo}{titleblue}{0.2}
\maxcell{1767}{428}{0.9}

\heatcell{1088}{555}{shadeone}{titleblue}{0.1}
\heatcell{1293}{555}{shadethree}{titleblue}{0.3}
\heatcell{1499}{555}{shadeeight}{white}{0.8}
\maxcell{1767}{555}{0.8}

\end{tikzpicture}}
        \vspace{2pt}
        {\small \textbf{(b)} Layer--prompt response patterns}
    \end{minipage}
    \caption{\textbf{Motivation for Layer--Prompt Profiles (LPP).} \textbf{(a)} Relationship between ID histogram entropy and near-OOD AUROC across candidate layer subsets on MIDOG using UniMedCLIP. The weak correlation suggests that minimizing ID entropy does not necessarily identify the most OOD-discriminative representations. \textbf{(b)} Two illustrative images exhibit identical layer-wise maximum similarity scores despite different layer--prompt response patterns. Reducing these responses to scalar confidence scores discards information preserved by the complete layer--prompt profile.}
    \label{fig:lpp_motivation}
\end{figure*}

\begin{figure}[t]
    \centering
    \includegraphics[width=\linewidth]{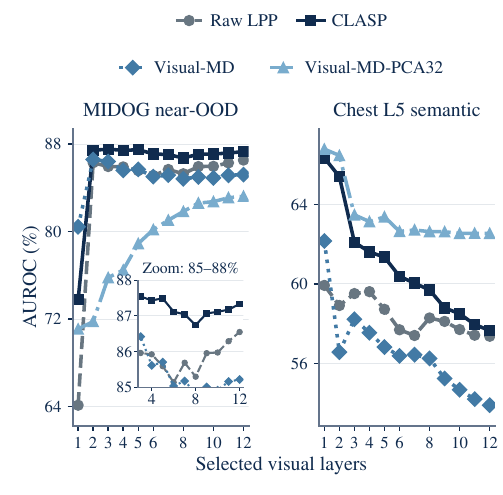}
    \caption{\textbf{Impact of layer selection and representation choice on OOD detection.} UniMedCLIP AUROC across selected-layer budgets for Chest X-ray L5 semantic OOD and MIDOG near-OOD. LPP stands for layer-prompt profiling and LPP-V combines LPP with visual representations.}
    \label{fig:lpp_layer_sweep}
\end{figure}

Existing VLM-based OOD detection methods typically rely on confidence scores derived from similarities between image representations and textual descriptions of known ID concepts. Maximum Concept Matching (MCM) \cite{ming2022mcm}, for instance, estimates confidence using the maximum softmax-normalized image--text similarity. While such approaches predominantly exploit final-layer representations, recent studies have investigated intermediate visual encoder layers and introduced entropy-guided strategies for selecting informative representations \cite{meza2025mysteries,rai2026multiresolution}. However, their effectiveness can vary substantially across distribution shifts and VLM backbones, with approaches that perform well in one setting often deteriorating in another. These limitations raise three fundamental questions concerning how to select complementary visual layers, preserve their prompt-conditioned information, and jointly exploit visual and textual representations for OOD detection.

\noindent\textbf{Q1: Can complementary visual layers be selected more effectively than through entropy minimization?} Existing entropy-guided approaches favor layers exhibiting concentrated confidence-score distributions on unlabeled ID samples. However, low ID entropy does not necessarily imply strong discrimination against unseen OOD samples. As illustrated in Fig.~\ref{fig:lpp_motivation}(a), our analysis reveals only a weak association between ID histogram entropy and near-OOD detection performance across candidate layer subsets. Beyond identifying informative layers, determining how many layers to retain is equally important. Fig.~\ref{fig:lpp_layer_sweep} illustrates that the preferred number of layers differs across distribution shifts: incorporating additional layers improves near-OOD detection on MIDOG before performance largely stabilizes, whereas fewer layers are preferable for Chest X-ray semantic OOD. These observations motivate a complementarity-aware selection strategy that automatically determines a compact set of visual layers without relying on entropy minimization or a predefined layer budget.

\noindent\textbf{Q2: Does scalar confidence aggregation discard useful layer--prompt information?} Conventional confidence-based approaches reduce image--text similarities to a single score, potentially discarding individual responses to different textual descriptions. As illustrated in Fig.~\ref{fig:lpp_motivation}(b), two images can exhibit identical layer-wise maximum similarities despite substantially different layer--prompt response patterns. Such differences may contain information relevant to identifying distribution shifts that cannot be recovered from the maximum scores alone. This motivates preserving the structured image--text similarity responses, or layer--prompt profiling (LPP) across selected visual layers and ID textual prompts, rather than collapsing them into scalar confidence estimates.

\noindent\textbf{Q3: Can visual and prompt-conditioned representations be jointly exploited for OOD detection?} Vision-language models provide two complementary sources of information: visual embeddings encode image characteristics, while image--text similarities capture their alignment with predefined semantic descriptions. Although layer--prompt profiles preserve structured semantic responses, they may not fully capture visual variations that are weakly reflected in the available textual prompts. Conversely, visual embeddings alone do not explicitly represent the relationships between images and ID concepts. This motivates jointly modeling visual features and matched layer--prompt profiles, allowing both representation types to contribute to a unified OOD detection framework across heterogeneous distribution shifts.

Motivated by these observations, we propose \textbf{CLASP (Complementary Layer Selection and Profiling)}, a zero-shot OOD detection framework built around three components. First, CLASP employs Gaussian conditional mutual information (CMI) to rank visual encoder layers according to their complementary information relative to an ID-fitted visual reference, followed by anchored spectral admission to automatically determine which layers to retain without a predefined layer budget. Second, the selected layers define a matched layer--prompt profile that preserves their individual image--text similarity responses. Third, CLASP combines this structured profile with compact visual representations from the same selected layers and models their joint distribution using a single multivariate Gaussian with Ledoit--Wolf shrinkage-regularized covariance, fitted exclusively on unlabeled in-distribution (ID) calibration images. During inference, squared Mahalanobis distance from the estimated ID distribution serves as the OOD anomaly score. CLASP requires no VLM fine-tuning, external OOD data, or test-time adaptation. \textbf{Our main contributions are as follows:}
\begin{itemize}
    \item We introduce a \textbf{complementarity-aware layer-selection strategy} that replaces entropy-guided ranking with Gaussian Visual CMI and employs anchored spectral admission to automatically determine a compact subset of visual encoder layers using only unlabeled ID data, without requiring a predefined layer budget.
    \item We propose \textbf{matched layer--prompt profiling}, preserving the structured image--text similarity responses of selected layers instead of reducing them to scalar confidence estimates. Our analysis demonstrates that these response patterns retain information that maximum confidence aggregation discards.
    \item We develop a \textbf{unified visual--prompt OOD detector} that jointly models matched layer--prompt profiles and compact visual representations using an ID-fitted Gaussian. Evaluations across histopathology, chest X-ray, and brain MRI benchmarks with two medical VLMs demonstrate competitive performance across heterogeneous shifts, while ablations and calibration stability experiments characterize the contributions and robustness of the framework.
\end{itemize}

\section{Related Work}
\label{sec:related_work}

\textbf{Out-of-Distribution Detection.} Conventional OOD detection methods rely on predictive confidence, feature statistics, or distances in learned representation spaces. Early approaches include Maximum Softmax Probability (MSP) \cite{hendrycks2017baseline}, ODIN \cite{liang2018odin}, and energy-based scoring \cite{liu2020energy}, while feature-based approaches employ Mahalanobis distance \cite{lee2018mahalanobis}, activation rectification (ReAct) \cite{sun2021react}, virtual-logit matching (ViM) \cite{wang2022vim}, and nearest-neighbor distances \cite{sun2022knn}. These methods typically operate on representations obtained from supervised classification models. Their effectiveness under medical distribution shifts has been investigated in OpenMIBOOD \cite{gutbrod2025openmibood}, highlighting the challenges of generalization across imaging domains.

\textbf{Zero-Shot OOD Detection with Vision--Language Models.} Pretrained VLMs such as CLIP \cite{radford2021clip} enable OOD detection by measuring image--text similarity against descriptions of known ID concepts, without task-specific classifier training. Maximum Concept Matching (MCM) \cite{ming2022mcm} estimates ID confidence from the maximum softmax-normalized similarity, while GL-MCM \cite{miyai2025glmcm} incorporates local visual information. However, confidence-based scoring may overlook unknown concepts that exhibit strong semantic similarity to the ID classes. To address this limitation, several approaches introduce negative or unknown concepts through generated descriptions or external textual knowledge, including ZOC \cite{esmaeilpour2022zoc}, NegLabel \cite{jiang2024neglabel}, and EOE \cite{cao2024eoe}. Other approaches, such as CLIPScope \cite{fu2025clipscope}, explore alternative scoring mechanisms to improve ID--OOD discrimination.

A complementary direction improves OOD detection through prompt learning and model adaptation. Building on prompt optimization approaches such as CoOp \cite{zhou2022coop}, methods including CLIPN \cite{wang2023clipn} and NegPrompt \cite{li2024negprompt} learn negative or negation-aware textual representations. Few-shot methods such as LoCoOp \cite{miyai2023locoop}, ID-like Prompt Learning \cite{bai2024idlike}, and Local-Prompt \cite{zeng2025localprompt} further exploit local information or synthetic outlier representations to improve detection. AdaNeg \cite{zhang2024adaneg} instead constructs adaptive negative proxies using test-time observations. While these methods improve semantic discrimination through additional concepts, learned prompts, or adaptation, they do not explicitly preserve the joint similarity responses across visual encoder layers and ID textual prompts.

\textbf{Medical VLM-Based OOD Detection.} Domain-specific models such as BiomedCLIP \cite{zhang2025biomedclip} and UniMed-CLIP \cite{khattak2024unimedclip} have extended vision--language learning to medical imaging. Ju et al. \cite{ju2025medicalood} investigated semantic and covariate shifts in medical VLMs and introduced hierarchical prompting. PROTON and PROTON+ \cite{das2026proton,das2026protonplus} combine image--text confidence with prototype-based distances using test-time memory updates.

\textbf{Intermediate Representations for OOD Detection.} Recent studies suggest that intermediate encoder layers contain complementary OOD-relevant information. Beyond VLMs, BLOOD \cite{jelenic2024blood} exploits between-layer transformation smoothness for OOD detection in transformers. Meza De la Jara et al. \cite{meza2025mysteries} investigated entropy-guided layer selection in CLIP, while Rai et al. \cite{rai2026multiresolution} introduced multi-resolution entropy estimation for medical VLMs. More recently, Voyager \cite{delajara2026voyager} proposed learned input-dependent layer routing using weak OOD supervision, whereas SeTAR \cite{li2024setar} modifies pretrained weight matrices through selective low-rank approximation. Unlike approaches that select, aggregate, or modify individual layer representations, LPP retains the complete layer--prompt similarity matrix and models its joint ID distribution using a single shrinkage-regularized Gaussian, with squared Mahalanobis distance providing the OOD anomaly score.

\section{Method}
\label{sec:method}

\begin{figure*}[t]
    \centering
    \includegraphics[width=\textwidth]{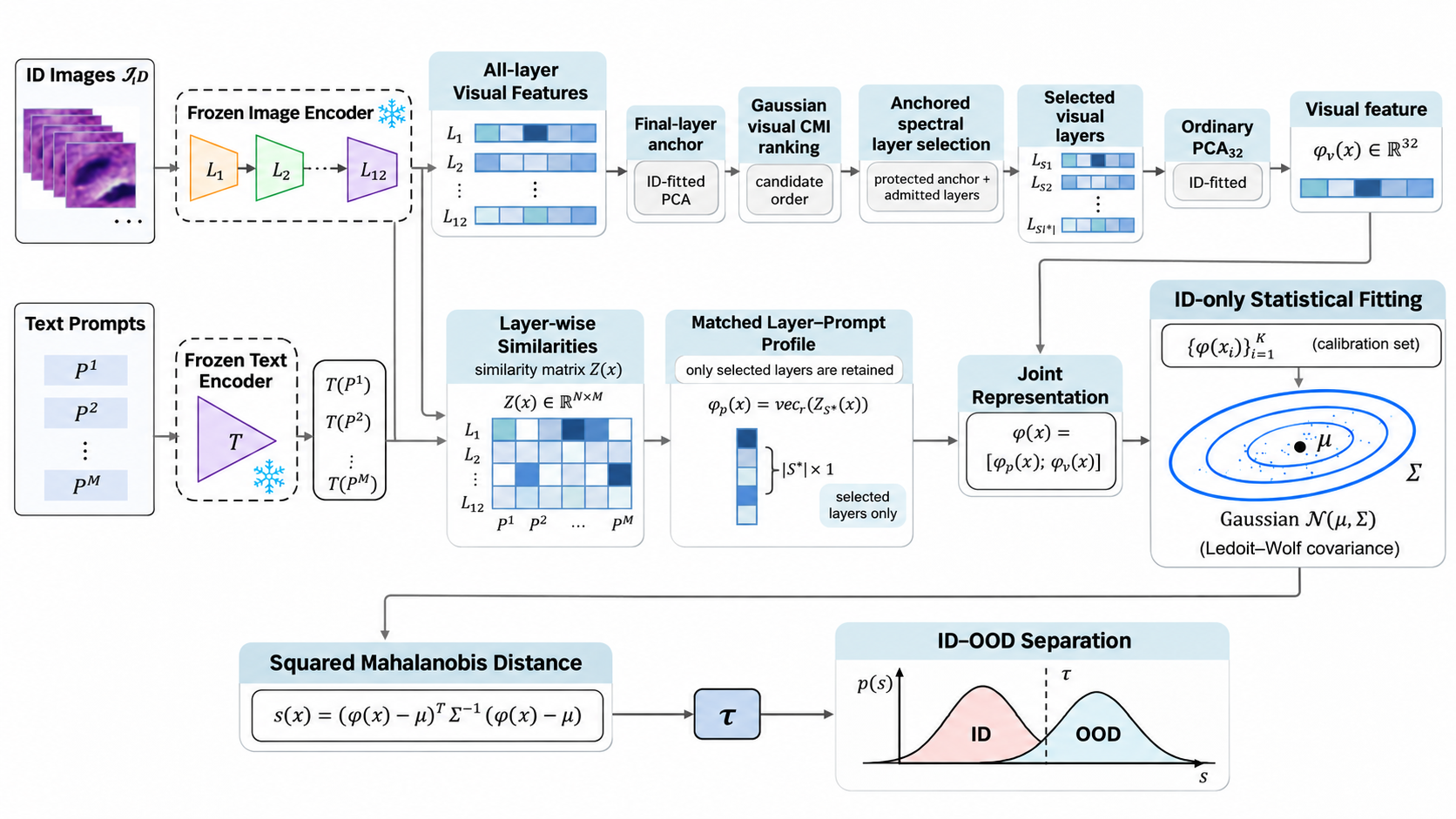}
    \caption{\textbf{Overview of CLASP.} A frozen VLM extracts visual embeddings and layer--prompt similarities. Gaussian conditional mutual information (CMI) ranks candidate layers, while anchored spectral admission determines how many ranked layers to retain within a fixed visual capacity. The admitted layers define a matched layer--prompt profile, which is jointly modeled with complementary visual features using an ID-fitted Gaussian. Squared Mahalanobis distance serves as the OOD anomaly score.}
    \label{fig:lpp_overview}
\end{figure*}

We propose \textbf{CLASP (Complementary Layer Selection and Profiling)}, a zero-shot OOD detection framework that combines CMI-guided layer ranking, anchored spectral admission, and matched visual--prompt modeling. Their prompt responses are combined with complementary visual features and modeled using a single Gaussian fitted exclusively on unlabeled ID calibration images. An overview is presented in Fig.~\ref{fig:lpp_overview}.

\subsection{Layer--Prompt Similarity Representation}
\label{sec:layer_prompt_similarity}

Let $\mathcal{D}_{\mathrm{ID}}=\{x_i\}_{i=1}^{N}$ denote the unlabeled ID calibration set and $\mathcal{P}=\{p_k\}_{k=1}^{K}$ the predefined ID textual prompts. Given a frozen VLM with visual and text encoders $f_v$ and $f_t$, we extract projected visual embeddings $\{\mathbf{v}_{\ell}(x)\}_{\ell=1}^{L}$ from all $L$ visual layers and text embeddings $\mathbf{t}_k=f_t(p_k)$. Their image--text similarity is defined as
\begin{equation}
    Z_{\ell k}(x)=\operatorname{sim}\left(\mathbf{v}_{\ell}(x),\mathbf{t}_k\right),
    \label{eq:lpp_similarity}
\end{equation}
where $\operatorname{sim}(\cdot,\cdot)$ follows the VLM-specific projection and normalization conventions. Collecting these responses yields the layer--prompt similarity matrix $\mathbf{Z}(x)\in\mathbb{R}^{L\times K}$, where each row $\mathbf{z}_{\ell}(x)\in\mathbb{R}^{K}$ retains the individual similarities of layer $\ell$ to all ID prompts. Unlike confidence-based approaches, CLASP preserves these responses without maximum aggregation.

\subsection{Layer Ranking and Selection}
\label{sec:layer_ranking_selection}

\noindent\textbf{CMI layer ranking.} We employ Gaussian conditional mutual information (CMI) to rank encoder layers according to the additional information their prompt responses provide about a visual reference. Let $\mathbf{a}(x)\in\mathbb{R}^{d_v}$ denote the final-layer visual embedding compressed using ID-fitted, unwhitened PCA, with $d_v=32$. Starting from the mandatory final-layer anchor $\mathcal{R}_1=\{L\}$, we evaluate each remaining candidate layer $\ell$ using
\begin{equation}
    \begin{aligned}
    \mathcal{I}(\ell\mid\mathcal{R})
    &=\frac{1}{K}I\left(\mathbf{z}_{\ell};\mathbf{a}\mid\mathbf{Z}_{\mathcal{R}}\right)\\
    &=\frac{1}{2K}\log\frac{\det\left(\boldsymbol{\Sigma}_{\ell\mid\mathcal{R}}\right)}{\det\left(\boldsymbol{\Sigma}_{\ell\mid\mathcal{R},a}\right)},
    \end{aligned}
    \label{eq:lpp_visual_cmi}
\end{equation}
where $\mathbf{Z}_{\mathcal{R}}$ concatenates prompt responses from previously ranked layers, and the conditional covariances are computed through Schur complements of an ID-fitted Ledoit--Wolf covariance. At each step, the highest-scoring candidate is appended to the ranking:
\begin{equation}
    \ell^{\star}=\underset{\ell\notin\mathcal{R}}{\arg\max}\;\mathcal{I}(\ell\mid\mathcal{R}),
    \qquad
    \mathcal{R}\leftarrow\mathcal{R}\cup\{\ell^{\star}\}.
    \label{eq:lpp_cmi_ranking}
\end{equation}
This greedy procedure produces a complete layer ranking conditioned on all previously ranked layers, which remains fixed during subsequent spectral admission. Importantly, CMI determines only the candidate ordering, not the final layer subset or its size.

\noindent\textbf{Anchored Spectral Layer Selection.} Given the frozen CMI ranking, we determine which layers to retain using an anchored spectral admission procedure with a fixed visual capacity of $d_v=32$. We first construct the final-layer visual anchor using ID-fitted, unwhitened PCA:
\begin{equation}
    \mathbf{a}(x)=\mathbf{U}_{L}^{\top}\left(\mathbf{v}_{L}(x)-\bar{\mathbf{v}}_{L}\right)\in\mathbb{R}^{d_v},
    \label{eq:lpp_anchor_pca}
\end{equation}
where $\mathbf{U}_{L}$ contains the leading ID principal directions. To identify empirically stable anchor coordinates, we compare the full-calibration covariance $\widehat{\boldsymbol{\Sigma}}_{a}$ with covariances estimated after deleting each calibration source group $g\in\mathcal{G}$. We define the perturbation radius as
\begin{equation}
    \tau_a=\max_{g\in\mathcal{G}}\left\|\widehat{\boldsymbol{\Sigma}}_{a}-\widehat{\boldsymbol{\Sigma}}_{a}^{(-g)}\right\|_{F}.
    \label{eq:lpp_anchor_radius}
\end{equation}
Anchor directions whose covariance eigenvalues exceed $\tau_a$ form the protected set $\mathcal{J}$, leaving $d_v-|\mathcal{J}|$ dimensions available for complementary visual directions.

For each candidate layer $\ell$ examined in CMI order, we obtain its ID-fitted PCA representation $\mathbf{b}_{\ell}(x)\in\mathbb{R}^{d_v}$ and remove the information linearly predictable from protected anchor coordinates and previously admitted visual residuals. Let $\boldsymbol{\psi}_{\mathcal{S}}(x)$ denote the previously admitted residual coordinates and $\mathbf{c}_{\mathcal{S}}(x)=[\mathbf{a}_{\mathcal{J}}(x);\boldsymbol{\psi}_{\mathcal{S}}(x)]$ their conditioning vector. The conditional visual residual is
\begin{equation}
    \begin{aligned}
    \mathbf{r}_{\ell}(x)
    &=\mathbf{b}_{\ell}(x)-\bar{\mathbf{b}}_{\ell}\\
    &\quad-\widehat{\boldsymbol{\Sigma}}_{bc}\widehat{\boldsymbol{\Sigma}}_{cc}^{-1}
    \left(\mathbf{c}_{\mathcal{S}}(x)-\bar{\mathbf{c}}_{\mathcal{S}}\right),
    \end{aligned}
    \label{eq:lpp_conditional_visual_residual}
\end{equation}
where the regression coefficients are estimated exclusively from unlabeled ID data. The residual covariance is eigendecomposed as $\widehat{\boldsymbol{\Sigma}}_{r_{\ell}}=\mathbf{Q}_{\ell}\operatorname{diag}(\boldsymbol{\gamma}_{\ell})\mathbf{Q}_{\ell}^{\top}$. Analogously to Eq.~\ref{eq:lpp_anchor_radius}, we estimate a source-deletion perturbation radius $\tau_{\ell}$ for the conditional residual covariance, re-estimating the conditional regression after each deletion while retaining the full-fit PCA bases and normalization conventions. Residual directions satisfying $\gamma_{\ell j}>\tau_{\ell}$ are eligible for admission, subject to the remaining visual capacity. A candidate layer is retained if it contributes at least one admitted direction. The procedure terminates when the first candidate contributes no admissible directions or the visual capacity is exhausted; any unused capacity is filled with remaining final-layer anchor coordinates. The resulting layer subset $\mathcal{S}^{\star}$ is therefore determined by spectral admission along the frozen CMI ranking, without requiring a predefined number of layers. The deletion criterion measures empirical stability under observed ID-source perturbations rather than guaranteeing OOD-optimal layer selection.

\subsection{Matched Layer--Prompt and Visual Representation}
\label{sec:lpp_representation}

Let $\mathcal{S}^{\star}$ denote the spectrally admitted layers, with $M=|\mathcal{S}^{\star}|$. We construct the matched layer--prompt representation by retaining all $K$ prompt responses from each admitted layer:
\begin{equation}
    \boldsymbol{\phi}_{p}(x)=\operatorname{vec}_{r}\left(\mathbf{Z}_{\mathcal{S}^{\star}}(x)\right)\in\mathbb{R}^{MK}.
    \label{eq:lpp_selected_profile}
\end{equation}
For the visual representation, we concatenate the normalized visual embeddings of the admitted layers in a fixed order, with the original selected-layer scaling, and apply unwhitened PCA fitted exclusively on ID calibration images. Denoting this concatenated representation by $\mathbf{v}_{\mathcal{S}^{\star}}(x)$, we obtain
\begin{equation}
    \boldsymbol{\phi}_{v}(x)=
    \mathbf{U}_{\mathcal{S}^{\star}}^{\top}
    \left(\mathbf{v}_{\mathcal{S}^{\star}}(x)-\bar{\mathbf{v}}_{\mathcal{S}^{\star}}\right)
    \in\mathbb{R}^{d_v},
    \qquad d_v=32,
    \label{eq:lpp_visual_representation}
\end{equation}
where $\mathbf{U}_{\mathcal{S}^{\star}}$ contains the leading ID principal directions. Thus, anchored spectral admission determines the retained layers, whereas ordinary PCA constructs their compact visual representation. The matched prompt and visual features are concatenated to form
\begin{equation}
    \mathbf{h}(x)=
    \begin{bmatrix}
        \boldsymbol{\phi}_{p}(x)\\
        \boldsymbol{\phi}_{v}(x)
    \end{bmatrix}
    \in\mathbb{R}^{D},\qquad D=MK+d_v.
    \label{eq:lpp_joint_representation}
\end{equation}
The resulting representation preserves the individual prompt responses of the admitted layers alongside their compact visual embeddings, which are jointly modeled using the ID-fitted Gaussian described in the following subsection.

\subsection{ID Distribution Modeling and OOD Scoring}
\label{sec:lpp_scoring}

We standardize each coordinate of $\mathbf{h}(x)$ using its ID calibration mean $\boldsymbol{\mu}_{h}$ and standard deviation $\boldsymbol{\sigma}_{h}$, obtaining $\widetilde{\mathbf{h}}(x)=(\mathbf{h}(x)-\boldsymbol{\mu}_{h})/(\boldsymbol{\sigma}_{h}+\epsilon)$. We then fit a single Gaussian $\mathcal{N}(\boldsymbol{\mu},\boldsymbol{\Sigma})$ to the standardized ID representations. Given their empirical covariance $\mathbf{S}$, we apply Ledoit--Wolf shrinkage~\cite{ledoit2004well} with numerical ridge regularization:
\begin{equation}
    \widehat{\boldsymbol{\Sigma}}=
    (1-\widehat{\rho})\mathbf{S}
    +\widehat{\rho}\frac{\operatorname{tr}(\mathbf{S})}{D}\mathbf{I}_{D}
    +\lambda\mathbf{I}_{D},
    \label{eq:lpp_covariance}
\end{equation}
where $\widehat{\rho}\in[0,1]$ is the shrinkage coefficient and $\lambda>0$ ensures numerical stability. For an unseen image $x$, the OOD anomaly score is its squared Mahalanobis distance from the learned ID distribution:
\begin{equation}
    \begin{aligned}
    s_{\mathrm{CLASP}}(x)
    &=
    \left(\widetilde{\mathbf{h}}(x)-\boldsymbol{\mu}\right)^{\top}\\
    &\quad\widehat{\boldsymbol{\Sigma}}^{-1}
    \left(\widetilde{\mathbf{h}}(x)-\boldsymbol{\mu}\right).
    \end{aligned}
    \label{eq:lpp_ood_score}
\end{equation}
Higher scores indicate greater deviations from the ID distribution. 


\begin{table*}[t]
\centering
\small
\setlength{\tabcolsep}{2pt}
\caption{MIDOG OOD detection: unweighted macro averages across seven near-OOD and two far-OOD domains. AUROC $\uparrow$ and FPR@95 $\downarrow$ are in \%. Explicit class-identity use denotes class-specific assignments or modeling beyond predefined ID text descriptions. Voyager is not reported because MIDOG contains only one semantic ID class, described by ten synonymous prompts, making its class-based layer-expert scores degenerate. PROTON+ results are averaged over five balanced online streams due to its test-time adaptation protocol, whereas the other methods are evaluated on complete test populations. Best results are highlighted in bold.}
\label{tab:midog_full_comparison}
\resizebox{0.94\textwidth}{!}{%
\begin{tabular}{lcccrrrrrrrr}
\toprule
\multirow{3}{*}{Method}
& \multirow{3}{*}{\shortstack{Explicit class-\\identity use}}
& \multirow{3}{*}{\shortstack{Ext.\\OOD}}
& \multirow{3}{*}{\shortstack{Additional\\Training}}
& \multicolumn{4}{c}{UniMedCLIP}
& \multicolumn{4}{c}{BioMedCLIP} \\
\cmidrule(lr){5-8}\cmidrule(lr){9-12}
& & & &
\multicolumn{2}{c}{Near-OOD}
& \multicolumn{2}{c}{Far-OOD}
& \multicolumn{2}{c}{Near-OOD}
& \multicolumn{2}{c}{Far-OOD} \\
\cmidrule(lr){5-6}\cmidrule(lr){7-8}
\cmidrule(lr){9-10}\cmidrule(lr){11-12}
& & & &
AUROC $\uparrow$ & FPR@95 $\downarrow$
& AUROC $\uparrow$ & FPR@95 $\downarrow$
& AUROC $\uparrow$ & FPR@95 $\downarrow$
& AUROC $\uparrow$ & FPR@95 $\downarrow$ \\
\midrule
MCM~\cite{ming2022mcm} & \xmark & \xmark & \xmark & 47.25 & 91.95 & 81.64 & 51.56 & 40.08 & 97.61 & 36.95 & 100.00 \\
SRE~\cite{meza2025mysteries} & \xmark & \xmark & \xmark & 29.10 & 99.70 & 97.10 & 16.60 & 32.10 & 97.30 & \textbf{99.80} & 1.30 \\
MRE~\cite{rai2026multiresolution} & \xmark & \xmark & \xmark & 54.60 & 94.40 & 97.40 & 13.40 & 36.40 & 100.00 & 99.50 & 2.70 \\
GL-MCM~\cite{miyai2025glmcm} & \xmark & \xmark & \xmark & 47.53 & 92.98 & 81.29 & 53.31 & 42.19 & 96.55 & 54.27 & 97.08 \\
PROTON+~\cite{das2026protonplus} & \cmark & \xmark & \xmark & 47.00 & 95.14 & 74.92 & 67.02 & 48.88 & 95.73 & 24.31 & 99.92 \\
Voyager~\cite{delajara2026voyager} & \cmark & \cmark & \cmark & -- & -- & -- & -- & -- & -- & -- & -- \\
V-Mahalanobis & \xmark & \xmark & \xmark & 80.48 & 47.01 & 99.99 & 0.10 & \textbf{72.97} & \textbf{62.24} & 99.67 & \textbf{1.07} \\
\rowcolor{lppRow}
CLASP (ours)
& \xmark & \xmark & \xmark
& \textbf{87.54} & \textbf{30.88}
& \textbf{100.00} & \textbf{0.00}
& 70.63 & 66.47
& 99.46 & 1.66 \\
\bottomrule
\end{tabular}%
}
\end{table*}

\begin{table*}[t]
\centering
\small
\caption{Chest X-ray and Brain MRI OOD detection under the L1 (2 prompts) and L5 (8 prompts) settings. Each cell reports AUROC $\uparrow$ / FPR@95 $\downarrow$ (in \%). The Brain MRI column reports the OASIS-3 versus BraTS-GLI benchmark using four ID prompts. Bold denotes the best reported result for each backbone and OOD category.}
\label{tab:chest_comparison}
\resizebox{0.97\textwidth}{!}{%
\begin{tabular}{llcccccc!{\hspace{6pt}\vrule width 0.5pt\hspace{6pt}}c}
\toprule
\multirow{2}{*}{Backbone} & \multirow{2}{*}{Method}
& \multicolumn{3}{c}{Chest X-ray: L1 (2 ID prompts)}
& \multicolumn{3}{c}{Chest X-ray: L5 (8 ID prompts)}
& Brain MRI \\
\cmidrule(lr){3-5}\cmidrule(lr){6-8}\cmidrule(l){9-9}
& & Semantic & Covariate & Far-OOD & Semantic & Covariate & Far-OOD & Near-OOD \\
\midrule
\multirow{8}{*}{UniMedCLIP}
& MCM~\cite{ming2022mcm} & 54.20/89.67 & 46.76/87.50 & 27.56/99.67 & 54.18/\textbf{77.50} & 78.43/46.50 & 84.62/43.00 & 38.90/100.00 \\
& SRE~\cite{meza2025mysteries} & 64.32/87.50 & 20.17/99.67 & 82.81/86.33 & 59.51/89.00 & 93.10/30.33 & \textbf{100.00}/\textbf{0.00} & 11.50/100.00 \\
& MRE~\cite{rai2026multiresolution} & 64.32/87.50 & 20.17/99.67 & 82.81/86.33 & 60.37/89.75 & 88.25/48.08 & \textbf{100.00}/\textbf{0.00} & 15.20/100.00 \\
& GL-MCM~\cite{miyai2025glmcm} & 55.36/92.83 & 47.71/91.00 & 33.18/99.33 & 54.46/79.33 & 81.18/55.33 & 91.52/34.00 & 40.40/95.00 \\
& PROTON+~\cite{das2026protonplus} & 61.39/88.13 & 93.88/29.00 & 91.52/37.93 & 49.90/92.13 & 88.50/36.53 & 96.86/9.53 & 35.90/100.00 \\
& Voyager~\cite{delajara2026voyager} & 52.42/92.83 & 25.29/98.17 & 31.08/98.33 & 52.63/88.50 & 72.47/74.67 & 74.33/71.17 & -- \\
& V-Mahalanobis & 62.00/86.67 & \textbf{99.97}/\textbf{0.00} & \textbf{100.00}/\textbf{0.00} & 62.00/86.67 & \textbf{99.97}/\textbf{0.00} & \textbf{100.00}/\textbf{0.00} & 51.60/92.50 \\
\rowcolor{lppRow}
& CLASP (ours) & \textbf{66.44}/\textbf{85.00} & 99.91/0.17 & \textbf{100.00}/\textbf{0.00} & \textbf{65.40}/87.83 & 99.89/\textbf{0.00} & \textbf{100.00}/\textbf{0.00} & \textbf{99.97}/\textbf{0.00} \\
\midrule
\multirow{8}{*}{BioMedCLIP}
& MCM~\cite{ming2022mcm} & 54.50/94.67 & 56.44/79.00 & 56.10/91.17 & 57.05/\textbf{91.50} & 54.50/89.33 & 69.58/71.33 & 56.30/90.00 \\
& SRE~\cite{meza2025mysteries} & 51.50/93.83 & 16.21/99.83 & 99.41/0.83 & 55.72/93.83 & 62.75/79.50 & 99.84/0.33 & 14.10/100.00 \\
& MRE~\cite{rai2026multiresolution} & 51.50/93.83 & 16.21/99.83 & 99.41/0.83 & 57.24/92.25 & 80.24/50.50 & 99.97/0.08 & 14.10/100.00 \\
& GL-MCM~\cite{miyai2025glmcm} & 55.79/93.50 & 54.39/82.33 & 51.05/93.17 & 54.90/92.83 & 50.42/92.83 & 76.18/57.50 & 57.90/95.00 \\
& PROTON+~\cite{das2026protonplus} & 55.57/93.93 & 75.79/67.63 & 91.81/15.27 & \textbf{58.00}/92.30 & 76.86/56.90 & 91.98/12.70 & 43.00/95.00 \\
& Voyager~\cite{delajara2026voyager} & 55.88/\textbf{91.00} & 63.16/75.00 & 17.86/99.83 & 57.81/92.00 & 70.07/80.00 & 19.38/100.00 & -- \\
& V-Mahalanobis & 53.16/94.83 & 95.67/\textbf{19.67} & \textbf{100.00}/\textbf{0.00} & 53.16/94.83 & 95.67/19.67 & \textbf{100.00}/\textbf{0.00} & 72.70/62.50 \\
\rowcolor{lppRow}
& CLASP (ours) & \textbf{57.34}/94.67 & \textbf{95.96}/20.67 & \textbf{100.00}/\textbf{0.00} & 56.75/94.17 & \textbf{97.29}/\textbf{14.33} & \textbf{100.00}/\textbf{0.00} & \textbf{100.00}/\textbf{0.00} \\
\bottomrule
\end{tabular}%
}
\end{table*}

\section{Experiments and Analysis}
\subsection{Datasets and Experimental Settings}
\label{sec:experimental_setup}

\noindent\textbf{Datasets.} We evaluate our approach on three medical imaging benchmarks spanning histopathology, chest X-ray, and brain MRI. For histopathology, we follow the MIDOG benchmark protocol of OpenMIBOOD \cite{gutbrod2025openmibood,rai2026multiresolution}, using domain 1a as ID with 1,815 unlabeled calibration images and 513 held-out ID images. Seven domains (2--5, 6a, 6b, and 7) are considered near-OOD, exhibiting semantic and acquisition-related shifts, while CCAgT and FNAC 2019 serve as far-OOD datasets. For chest X-ray evaluation, we follow the radiology benchmark protocol of PROTON+ \cite{das2026protonplus}, using RSNA normal and pneumonia images as ID, NIH ChestX-ray14 images with other findings as semantic OOD, pediatric chest radiographs as covariate OOD, and Imagenette as far-OOD. The benchmark construction explicitly de-duplicates NIH images against RSNA to prevent overlap between ID and semantic-OOD samples. For brain MRI, we evaluate OASIS-3 versus BraTS-GLI, using healthy brain MR images from OASIS-3 as ID and glioma-containing MR volumes from BraTS-GLI as near-OOD. The MRI benchmark comprises 848 ID calibration volumes, 181 held-out ID volumes, and 1,125 OOD volumes.

\noindent\textbf{Experimental Setup.} We employ UniMedCLIP \cite{khattak2024unimedclip} and BioMedCLIP \cite{zhang2025biomedclip}, keeping both visual and textual encoders frozen. For the volumetric MRI benchmark, representations are aggregated across nine axial slices per volume. We report AUROC ($\uparrow$) and FPR@95 ($\downarrow$), with OOD as the positive class. MIDOG results are reported as unweighted macro averages across the seven near-OOD and two far-OOD domains. All feature normalization statistics, PCA projections, Gaussian parameters, and covariance regularization are estimated exclusively from unlabeled ID calibration samples and remain fixed during inference.

\subsection{Main Results}
\label{sec:main_results}

Tables~\ref{tab:midog_full_comparison} and~\ref{tab:chest_comparison} show that CLASP substantially outperforms confidence-based methods on MIDOG near-OOD, particularly with UniMedCLIP, while V-Mahalanobis remains competitive and performs better with BioMedCLIP. On Chest X-ray, CLASP achieves strong covariate- and far-OOD detection, whereas semantic shifts remain considerably more challenging. The results also demonstrate near-perfect discrimination between OASIS-3 and BraTS-GLI across both backbones. 
Overall, although individual baselines can outperform CLASP under specific distribution shifts or VLM backbones, CLASP maintains competitive performance across the evaluated settings without OOD-specific adaptation or method selection, suggesting a more broadly applicable detection framework rather than one optimized for a particular shift.

\subsection{Analysis and Ablations}
\label{sec:analysis_ablations}

\begin{table}[t]
\centering
\footnotesize
\setlength{\tabcolsep}{3pt}
\caption{Ablation of layer-ranking criteria using UniMedCLIP. Each ranking is followed by the same spectral admission. Results report AUROC $\uparrow$ / FPR@95 $\downarrow$ (in \%).}
\label{tab:ranking_ablation}
\resizebox{\columnwidth}{!}{%
\begin{tabular}{lcccc}
\toprule
\multirow{2}{*}{Ranking criterion}
& \multicolumn{2}{c}{MIDOG near}
& \multicolumn{2}{c}{Chest L5 semantic} \\
\cmidrule(lr){2-3}\cmidrule(lr){4-5}
& \shortstack{Selected\\layers} & AUROC/FPR@95
& \shortstack{Selected\\layers} & AUROC/FPR@95 \\
\midrule

\multicolumn{1}{l}{\shortstack[l]{Gaussian Visual\\CMI (ours)}}
& $\{3,8,12\}$ & \textbf{87.54}/\textbf{30.88}
& $\{4,12\}$ & 65.40/\textbf{87.83} \\

Entropy ordering
& $\{7,12\}$ & 80.31/51.43
& $\{11,12\}$ & 63.24/88.33 \\

\multicolumn{1}{l}{\shortstack[l]{Conditional\\ covariance}}
& $\{5,8,12\}$ & 85.14/36.73
& $\{5,12\}$ & \textbf{65.54}/88.17 \\

Marginal visual MI
& $\{11,12\}$ & 78.70/55.67
& $\{11,12\}$ & 63.24/88.33 \\

\bottomrule
\end{tabular}%
}
\end{table}

\noindent\textbf{Layer-ranking ablation.} Table~\ref{tab:ranking_ablation} that Gaussian Visual CMI provides an effective ordering of complementary layers, particularly on MIDOG. The results show that typically a combination of last layer along with intermediate layers gives better OOD detection performance, while the resulting performance remains dependent on the ranking criterion.

\begin{table}[t]
\centering
\footnotesize
\setlength{\tabcolsep}{4pt}
\caption{Ablation of covariance modeling using UniMedCLIP. All variants use identical CLASP representations. Results report AUROC/FPR@95 (\%).}
\label{tab:covariance_ablation}
\resizebox{0.82\columnwidth}{!}{%
\begin{tabular}{lcc}
\toprule
Covariance model & MIDOG near & Chest L5 semantic \\
\midrule
\shortstack[l]{Full Ledoit--Wolf\\covariance (ours)}
& \textbf{87.54}/\textbf{30.88}
& \textbf{65.40}/\textbf{87.83} \\

Diagonal covariance
& 80.16/49.82
& 58.24/92.83 \\

Identity covariance
& 80.16/49.82
& 58.24/92.83 \\
\bottomrule
\end{tabular}%
}
\end{table}

\noindent\textbf{Covariance modeling ablation.} Table~\ref{tab:covariance_ablation} demonstrates the importance of modeling cross-feature dependencies. Full Ledoit--Wolf covariance achieves the best performance on both benchmarks, improving AUROC over diagonal covariance by 7.38 percentage points on MIDOG and 7.15 points on Chest L5. Diagonal and identity covariance yield identical results, highlighting the importance of cross-coordinate correlations in the joint visual--prompt representation.

\begin{figure}[t]
    \centering
    \includegraphics[width=\columnwidth]{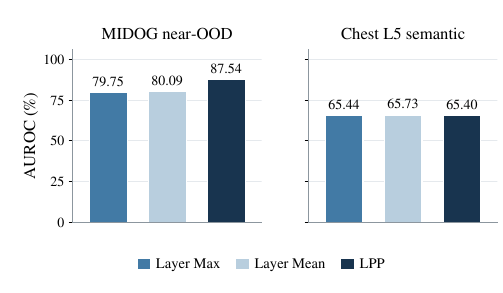}
    \caption{\textbf{Effect of layer--prompt aggregation.} Comparison of layer-wise maximum, mean, and complete layer--prompt profiling (LPP) using UniMedCLIP.}
    \label{fig:lpp_aggregation_ablation}
\end{figure}

\noindent\textbf{Layer--prompt aggregation.} Fig.~\ref{fig:lpp_aggregation_ablation} shows that preserving the complete layer--prompt profile substantially improves MIDOG near-OOD AUROC, achieving 87.54\% compared with 79.75\% and 80.09\% for layer-wise maximum and mean aggregation, respectively. On Chest L5 semantic OOD, all three approaches perform similarly, with mean aggregation achieving a marginally higher. These results suggest that preserving individual prompt responses is particularly beneficial for histopathology domain shifts, while its advantage is less pronounced for Chest X-ray semantic shifts.

\begin{figure}[t]
    \centering
    \includegraphics[width=\columnwidth]{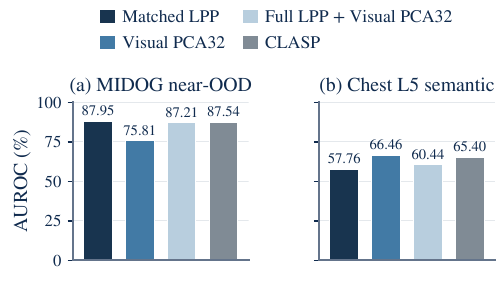}
    \caption{\textbf{Ablation of matched profiling and joint representation.} Comparison of matched LPP, visual PCA32, full LPP with visual PCA32, and CLASP with anchored spectral admission selected layers.}
    \label{fig:matched_joint_ablation}
\end{figure}

\noindent\textbf{Matched profiling and joint representation.} Fig.~\ref{fig:matched_joint_ablation} shows that matched profiling improves the joint representation over full LPP on both benchmarks, increasing AUROC from 87.21\% to 87.54\% on MIDOG and from 60.44\% to 65.40\% on Chest L5. However, the individual branches exhibit shift-dependent behavior: matched LPP performs best on MIDOG (87.95\%), characterized by acquisition and cell-type shifts, whereas visual PCA32 performs best on Chest L5 (66.46\%), involving previously unseen pathological findings. These findings highlight the complementary but shift-dependent contributions of prompt and visual representations.

\begin{table}[t]
\centering
\footnotesize
\setlength{\tabcolsep}{2pt}
\caption{ID calibration stability of CLASP using UniMedCLIP across ten repetitions with approximately 80\% calibration data. Agreement measures recovery of the full-calibration layer set, while Jaccard measures pairwise overlap between selected sets. AUROC and FPR@95 are reported as mean $\pm$ standard deviation (\%).}
\label{tab:calibration_stability}
\resizebox{\columnwidth}{!}{%
\begin{tabular}{lcccc}
\toprule
Benchmark & \shortstack{Reference\\layers}
& \shortstack{Agreement\\(\%)}
& Jaccard
& AUROC/FPR@95 \\
\midrule
MIDOG near
& $\{3,8,12\}$
& 50.00
& 0.72
& $86.74\pm0.64$ / $32.41\pm1.76$ \\

\shortstack{Chest L5\\semantic}
& $\{4,12\}$
& 30.00
& 0.54
& $65.02\pm0.84$ / $88.20\pm0.97$ \\
\bottomrule
\end{tabular}%
}
\end{table}

\noindent\textbf{ID calibration stability.} Table~\ref{tab:calibration_stability} evaluates CLASP under reduced ID calibration data. Although the exact selected layers vary across repetitions, the number of retained layers remains unchanged, with three layers on MIDOG and two on Chest L5. Despite moderate selection agreement, AUROC exhibits limited variation across repetitions, suggesting that CLASP's detection performance is less sensitive to calibration changes than the identities of individual selected layers. This indicates robustness in downstream detection while highlighting variability in the automatic layer selection.

\section{Conclusion}
\label{sec:conclusion}

We presented CLASP, a zero-shot OOD detection framework that combines Gaussian conditional mutual information for layer ranking, anchored spectral selection, and joint modeling of matched layer--prompt profiles and visual representations. By leveraging unlabeled ID calibration data, CLASP automatically determines a compact set of complementary layers and models their joint distribution using a single shrinkage-regularized Gaussian. Experiments across three medical imaging benchmarks and two VLM backbones demonstrate competitive and often improved OOD detection, particularly under acquisition-related distribution shifts, while highlighting the importance of preserving structured layer--prompt information and cross-feature dependencies.

\noindent\textbf{Limitations and future work.} CLASP's effectiveness remains dependent on the distribution shift and VLM backbone, with semantic OOD detection posing particular challenges. Furthermore, although detection performance remains relatively stable under calibration subsampling, the identities of selected layers can vary. Future work will investigate more robust layer-selection strategies and adaptive integration of visual and prompt representations.


{
    \small
    \bibliographystyle{ieeenat_fullname}
    \bibliography{main}
}


\end{document}